# Enriched Physics-informed Neural Networks for Dynamic Poisson-Nernst-Planck Systems

Xujia Huang[1,2], Fajie Wang[1,*], Benrong Zhang[1,2], Hanqing Liu[1,2]

[1] *College of Mechanical and Electrical Engineering, National Engineering Research Center for Intelligent Electrical Vehicle Power System, Qingdao University, Qingdao 266071, China*

[2] *School of Mathematics and Statistics, Qingdao University, Qingdao 266071, China*

## ABSTRACT

This paper proposes a meshless deep learning algorithm, enriched physics-informed neural networks (EPINNs), to solve dynamic Poisson-Nernst-Planck (PNP) equations with strong coupling and nonlinear characteristics. The EPINNs takes the traditional physics-informed neural networks as the foundation framework, and adds the adaptive loss weight to balance the loss functions, which automatically assigns the weights of losses by updating the parameters in each iteration based on the maximum likelihood estimate. The resampling strategy is employed in the EPINNs to accelerate the convergence of loss function. Meanwhile, the GPU parallel computing technique is adopted to accelerate the solving process. Four examples are provided to demonstrate the validity and effectiveness of the proposed method. Numerical results indicate that the new method has better applicability than traditional numerical methods in solving such coupled nonlinear systems. More importantly, the EPINNs is more accurate, stable, and fast than the traditional physics-informed neural networks. This work provides a simple and high-performance numerical tool for addressing PNPs with arbitrary boundary shapes and boundary conditions.

## 1. Introduction

The study of electrokinetic and transport phenomena in complex systems is critical in various fields such as biological, chemical engineering, material science, and electrochemistry. The transport of ions and molecules in these systems is significant for designing efficient drug delivery systems, optimising battery performance, and developing new biomaterials. To describe the electrodiffusion of mobile ions and charged ligands, the Poisson-Nernst-Planck (PNP) system [1, 2] is introduced as a standard model, which is a set of coupled partial differential equations (PDEs) consisted of the Poisson equation and the Nernst-Planck equation. The Poisson equation describes the relationship between the charge density and the electric potential. The Nernst-Planck equations depicts the mass transfer process of electric fluid in an electric field environment [3, 4].

[*] Corresponding author, E-mail address: wfj1218@126.com (Fajie Wang).

In the past few years, there have been lots of investigations into the PNP systems, especially in its numerical simulation. The finite difference method (FDM) [5, 6] with a rigorous error analysis has been widely utilized in solving various PNP systems. The finite element method (FEM) [7, 8] is another commonly used method. Numerous commercial software, such as COMSOL and ANSYS FLUENT, have been available for simulating PNP systems. In addition, the Chebyshev collocation techniques [9, 10] have also been applied to PNP systems. These approaches definitely have been both effective and fully play their respective advantages in addressing such problems. However, they also have some limitations that need to be further addressed, such as the troublesome grid generation in grid-dependent methods, as well as the inaccuracy numerically for strongly coupled nonlinear systems.

In recent years, the physics-informed neural networks (PINNs) has attracted growing attention due to its potential applications in solving nonlinear PDEs. Wen et al. [11] constructed the multiple parallel subnets of PINNs to solve the coupled nonlinear Schrodinger model and obtained good convergence results. Wu et al. [12] used the modified PINNs to predict the dynamics of optical pulses including one-soliton, two-soliton, and rogue waves based on the coupled nonlinear equation in birefringent fibres. Chang et al. [13] proposed a hybrid numerical method combining both conventional methods and the PINNs for solving the Ampère-Nernst-Planck equations. For an overview of the state of the art in existing theories, algorithms, and software packages, see Refs. [14-16].

This paper proposes a novel meshless deep learning algorithm, named enriched physics-informed neural networks (EPINNs), to solve the PNP problems. Unlike the traditional PINNs [17-23], the proposed method employs both the adaptive loss weight method [24] and the resampling strategy [25] to avoid inaccuracy in optimization affected by unbalanced gradients. The adaptive loss weight method automatically updates loss weights at the beginning of each iteration of training of each network. The resampling strategy is to change the data distribution on the training sample so as to make the model learn the characteristics of each class in a more balanced manner. Furthermore, the GPU parallel computing technique is adopted to accelerate the training process.

The rest of this paper is organized as follows. In Sec. 2 we introduced the PNP equations. Then we introduced the mathematical theory of the original PINNs and EPINNs in Sec. 3. In Sec. 4 we present our numerical experiment process and results. Finally, the paper ends with a summary in Sec. 5.

## 2. Poisson-Nernst-Planck equations

The PNP system can describe the evolution of positively and negatively charged particles (or ions), the classical unsteady dimensionless PNP system [26] is given by:

$$\begin{cases} \dfrac{\partial c_p}{\partial t} = \nabla \cdot (\nabla c_p + c_p \nabla \phi), & \text{in } \Omega_T := \Omega \times [0,T], \\ \dfrac{\partial c_n}{\partial t} = \nabla \cdot (\nabla c_n - c_n \nabla \phi), & \text{in } \Omega_T, \\ -\Delta \phi = c_p - c_n, & \text{in } \Omega_T, \end{cases} \quad (1)$$

where $c_p$ and $c_n$ represent the concentration of cations and anions, respectively, $\phi$ is the electric potential, $\Omega$ is a bounded domain, and $[0,T]$ is the time interval. In the above non-dimensional equations, the characteristic length scale is chosen as the Debye length and the characteristic time scale is chosen as the diffusive time scale.

In the present study, we considered the Dirichlet and Neumann boundary conditions. The Dirichlet boundary condition is given by:

$$\begin{cases} c_p(\mathbf{x},t) = \Gamma_1(\mathbf{x},t), & \mathbf{x} \in \partial\Omega, t \in [0,T], \\ c_n(\mathbf{x},t) = \Gamma_2(\mathbf{x},t), & \mathbf{x} \in \partial\Omega, t \in [0,T], \\ \phi(\mathbf{x},t) = \Gamma_3(\mathbf{x},t), & \mathbf{x} \in \partial\Omega, t \in [0,T]. \end{cases} \quad (2)$$

The Neumann boundary condition is given by:

$$\begin{cases} \dfrac{\partial c_p(\mathbf{x},t)}{\partial n} = N_1(\mathbf{x},t), & \mathbf{x} \in \partial\Omega, t \in [0,T], \\ \dfrac{\partial c_n(\mathbf{x},t)}{\partial n} = N_2(\mathbf{x},t), & \mathbf{x} \in \partial\Omega, t \in [0,T], \\ \dfrac{\partial \phi(\mathbf{x},t)}{\partial n} = N_3(\mathbf{x},t), & \mathbf{x} \in \partial\Omega, t \in [0,T], \end{cases} \quad (3)$$

where $\Gamma_1(\mathbf{x},t)$, $\Gamma_2(\mathbf{x},t)$, $\Gamma_3(\mathbf{x},t)$, $N_1(\mathbf{x},t)$, $N_2(\mathbf{x},t)$ and $N_3(\mathbf{x},t)$ are known functions, $\frac{\partial}{\partial n}$ represents the normal derivative of a physical variable, $\partial\Omega$ is the boundary of domain $\Omega$. In addition, we assume that the system is restricted to the following initial conditions:

$$\begin{cases} c_p(\mathbf{x},t_0) = p_0(\mathbf{x}), & \mathbf{x} \in \Omega, \\ c_n(\mathbf{x},t_0) = n_0(\mathbf{x}), & \mathbf{x} \in \Omega, \\ \phi(\mathbf{x},t_0) = \Lambda_0(\mathbf{x}), & \mathbf{x} \in \Omega, \end{cases} \quad (4)$$

where $p_0(\mathbf{x})$, $n_0(\mathbf{x})$ and $\Lambda_0(\mathbf{x})$ are known functions, $t_0$ is the initial time.

## 3. Numerical method

### 3.1 The framework of PINNs-based PNP system

To handle PNP problems described in Eqs. (1)-(4), we construct multilayer forward neural networks with $M$ layers to separately approximate the target solutions of PNP system. In the PINNs framework, the inputs of the neural network are the coordinates of the training points which consist of four parts: the training data $\{\mathbf{x}_f^j, t_f^j\}_{j=1}^{N_f}$ inside the computational domain $\Omega$, the training data $\{\mathbf{x}_b^j, t_b^j\}_{j=1}^{N_b}$ on the boundary, the training data $\{\mathbf{x}_i^j, t_i^j\}_{j=1}^{N_i}$ at the initial time, the training data $\{\mathbf{x}_{data}^j, t_{data}^j\}_{j=1}^{N_{data}}$ in the true data set, $N_0$, $N_b$, $N_f$ and $N_{data}$ are the number of initial training data, boundary training data, collocation points and true training data, $N_t$ is the total training data set. In the following equations, all partial derivatives are obtained via the automatic differentiation (AD) [27]. The PINNs integrates the prior knowledge of physics into the loss function to enhance the information content of the data. In the traditional PINNs, the loss function composed of eight loss terms is formulated as:

$$\begin{aligned} L_{total}(\theta; N_t) = &\lambda_f L_f(\theta; N_f) + \lambda_i L_{IC}(\theta; N_i) + \lambda_{b_1} L_{BC_1}(\theta; N_b) + \lambda_{b_2} L_{BC_2}(\theta; N_b) + \lambda_{b_3} L_{BC_3}(\theta; N_b) \\ &+ \lambda_{d_1} L_{data_1}(\theta; N_{data}) + \lambda_{d_2} L_{data_2}(\theta; N_{data}) + \lambda_{d_3} L_{data_3}(\theta; N_{data}), \end{aligned} \quad (5)$$

where $\lambda = \{\lambda_f, \lambda_i, \lambda_{b_1}, \lambda_{b_2}, \lambda_{b_3}, \lambda_{d_1}, \lambda_{d_2}, \lambda_{d_3}\}$ is the loss weight parameter, $L_f$ denotes the loss of PDEs, $L_{IC}$ denotes the loss of initial condition, $L_{BC_i}$ ($i=1,2,3$) denote the losses of boundary conditions, and $L_{data_i}$ ($i=1,2,3$) represent the differences between the true solutions and the neural network approximations at training data $\{\mathbf{x}_{data}^j, t_{data}^j\}_{j=1}^{N_{data}}$. Considering the Dirichlet boundary conditions and assuming the approximate solutions output by the neural network are $\hat{c}_p$, $\hat{c}_n$ and $\hat{\phi}$, the sub-losses in Eq. (5) can be represented as follows:

$$L_f(\theta; N_f) = \frac{1}{N_f} \sum_{j=1}^{N_f} \begin{pmatrix} \left| \frac{\partial \hat{c}_p(\mathbf{x}_f^j, t_f^j)}{\partial t} - \nabla \cdot (\nabla \hat{c}_p(\mathbf{x}_f^j, t_f^j) + \hat{c}_p(\mathbf{x}_f^j, t_f^j) \nabla \hat{\phi}(\mathbf{x}_f^j, t_f^j)) \right|^2 \\ + \left| \frac{\partial \hat{c}_n(\mathbf{x}_f^j, t_f^j)}{\partial t} - \nabla \cdot (\nabla \hat{c}_n(\mathbf{x}_f^j, t_f^j) + \hat{c}_n(\mathbf{x}_f^j, t_f^j) \nabla \hat{\phi}(\mathbf{x}_f^j, t_f^j)) \right|^2 \\ \left| -\Delta \hat{\phi}(\mathbf{x}_f^j, t_f^j) - \hat{c}_p(\mathbf{x}_f^j, t_f^j) + \hat{c}_n(\mathbf{x}_f^j, t_f^j) \right|^2 \end{pmatrix}, \quad (6)$$

$$L_{IC}(\theta; N_i) = \frac{1}{N_i} \sum_{j=1}^{N_0} \left( \left| \hat{c}_p(\mathbf{x}_i^j, t_0) - p_0(\mathbf{x}_i^j) \right|^2 + \left| \hat{c}_n(\mathbf{x}_i^j, t_0) - n_0(\mathbf{x}_i^j) \right|^2 + \left| \hat{c}_p(\mathbf{x}_i^j, t_0) - \Lambda_0(\mathbf{x}_i^j) \right|^2 \right), \quad (7)$$

$$L_{data_1}(\theta; N_{data}) = \frac{1}{N_{data}} \sum_{j=1}^{N_{data}} \left( \left| \hat{c}_p(\mathbf{x}_{data}^j, t_{data}^j) - c_p(\mathbf{x}_{data}^j, t_{data}^j) \right|^2 \right), \quad (8)$$

$$L_{data_2}(\theta; N_{data}) = \frac{1}{N_{data}} \sum_{j=1}^{N_{data}} \left( \left| \hat{c}_n(\mathbf{x}_{data}^j, t_{data}^j) - c_n(\mathbf{x}_{data}^j, t_{data}^j) \right|^2 \right), \quad (9)$$

$$L_{data_3}(\theta; N_{data}) = \frac{1}{N_{data}} \sum_{j=1}^{N_{data}} \left( \left| \hat{\phi}(\mathbf{x}_{data}^j, t_{data}^j) - \phi(\mathbf{x}_{data}^j, t_{data}^j) \right|^2 \right), \quad (10)$$

$$L_{BC_1}(\theta; N_b) = \frac{1}{N_b} \sum_{j=1}^{N_b} \left( \left| \hat{c}_p(\mathbf{x}_b^j, t_b^j) - \Gamma_1(\mathbf{x}_b^j, t_b^j) \right|^2 \right), \quad (11)$$

$$L_{BC_2}(\theta; N_b) = \frac{1}{N_b} \sum_{j=1}^{N_b} \left( \left| \hat{c}_n(\mathbf{x}_b^j, t_b^j) - \Gamma_2(\mathbf{x}_b^j, t_b^j) \right|^2 \right), \quad (12)$$

$$L_{BC_3}(\theta; N_b) = \frac{1}{N_b} \sum_{j=1}^{N_b} \left( \left| \hat{\phi}(\mathbf{x}_b^j, t_b^j) - \Gamma_3(\mathbf{x}_b^j, t_b^j) \right|^2 \right). \quad (13)$$

When we consider the Neumann boundary conditions, the loss terms $L_{BC_1}$, $L_{BC_2}$ and $L_{BC_3}$ should be replaced by:

$$L_{BC_1}(\theta; N_b) = \frac{1}{N_b} \sum_{j=1}^{N_b} \left( \left| \frac{\partial \hat{c}_p(\mathbf{x}_b^j, t_b^j)}{\partial n} - N_1(\mathbf{x}_b^j, t_b^j) \right|^2 \right), \quad (14)$$

$$L_{BC_2}(\theta; N_b) = \frac{1}{N_b} \sum_{j=1}^{N_b} \left( \left| \frac{\partial \hat{c}_n\left(\mathbf{x}_b^j, t_b^j\right)}{\partial n} - N_2\left(\mathbf{x}_b^j, t_b^j\right) \right|^2 \right), \tag{15}$$

$$L_{BC_3}(\theta; N_b) = \frac{1}{N_b} \sum_{j=1}^{N_b} \left( \left| \frac{\partial \hat{\phi}\left(\mathbf{x}_b^j, t_b^j\right)}{\partial n} - N_3\left(\mathbf{x}_b^j, t_b^j\right) \right|^2 \right). \tag{16}$$

*3.2 Enriched PINNs (EPINNs) by adaptive loss weights method and resampling strategy*

There is uncertainty in the prediction of PINNs due to the randomness of data or the limitations of the model architecture. To deal with the PNP problem accurately and effectively, the resampling strategy and adaptive loss weights extended from the work in Refs. [24, 25] are utilized for balancing multiple loss terms. The primary concept of the self-adaptive loss balancing method, as presented in Ref. [28], involves assigning weights to multiple loss functions by maximizing the Gaussian likelihood of uncertainty in a multi-task deep learning problem. In the context of the investigated PNP system in this paper, we define the following loss function:

$$\begin{aligned}L_{total}(\varepsilon; \theta; N_t) =\ & \frac{1}{2\varepsilon_f} L_f(\theta; N_f) + \frac{1}{2\varepsilon_i} L_{IC}(\theta; N_i) + \frac{1}{2\varepsilon_{b_1}} L_{BC_1}(\theta; N_b) + \frac{1}{2\varepsilon_{b_2}} L_{BC_2}(\theta; N_b) \\ & + \frac{1}{2\varepsilon_{b_3}} L_{BC_3}(\theta; N_b) + \frac{1}{2\varepsilon_{d_1}} L_{data_1}(\theta; N_{data}) + \frac{1}{2\varepsilon_{d_2}} L_{data_2}(\theta; N_{data}) \\ & + \frac{1}{2\varepsilon_{d_3}} L_{data_3}(\theta; N_{data}) + \log \varepsilon_f \varepsilon_i \varepsilon_{b_1} \varepsilon_{b_2} \varepsilon_{b_3} \varepsilon_{d_1} \varepsilon_{d_2} \varepsilon_{d_3}, \end{aligned} \tag{17}$$

where $\varepsilon = \{\varepsilon_f, \varepsilon_i, \varepsilon_{b_1}, \varepsilon_{b_2}, \varepsilon_{b_3}, \varepsilon_{d_1}, \varepsilon_{d_2}, \varepsilon_{d_3}\}$ describes the adaptive weights of loss terms. The goal of EPINNs is to find the best model weights $\theta^*$ and adaptive weights $\varepsilon^*$ by minimizing the loss $L_{total}(\varepsilon; \theta; N_t)$. For example, when the adaptive weight $\varepsilon_f$ decreases, the total weight $\frac{1}{2\varepsilon_f}$ increases, which makes the $L_f$ more punishing. Alternatively, the final term $\log \varepsilon_f$ prevents the adaptive weight $\varepsilon_f$ from decreasing excessively. A large adaptive weight will typically minimize the contribution of the loss term, whereas a small adaptive weight will increase its contribution and punish the model. This elucidates the automated adjustment mechanism for the adaptive weight associated with each loss component.

During the training process, we sample the collocation points from the spatial-temporal domain arbitrarily and dynamically. Ref. [25] indicates that resampling all training data in each iteration could reduce the stability and precision of training. Therefore, we introduce a resampling strategy, which resampling the training data at every $K\omega$ iterations, where $\omega$ is resampling ratio, $K$ is the total iterations. Assuming a resampling number of $n$ and iterating $K$ times with the Adam optimizer, the resampling ratio is given by $\omega = \frac{n}{K}$. Fig. 1 depicts a representation of EPINNs. Fig. 1 depicts the whole framework of EPINNs. According to the above procedure, the specific implementation steps of the proposed method are formulated in Algorithm 1.

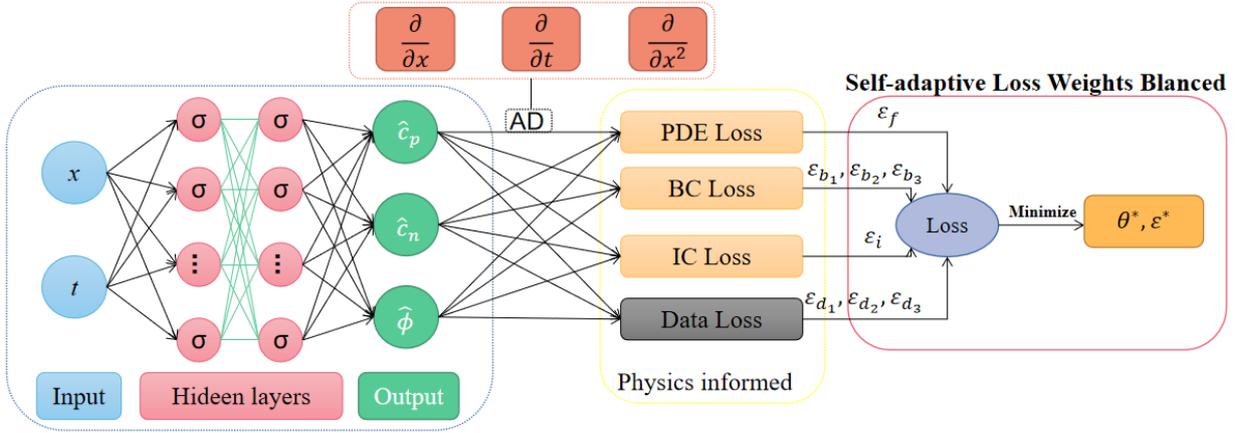

**Fig. 1.** The framework of EPINNs for solving the PNP equations.

**Algorithm 1:** EPINNs by adaptive loss weights method and resampling strategy

**Step1**: Consider a fully connected neural network to define the frame of EPINNs, training set $N_t = \{N_f, N_i, N_b, N_d\}$, sampling ratio set to $\omega$.

**Step2**: Compute the PDEs residual using AD, and define the loss terms.

**Step3**: Initialize the adaptive weight collection:

$$\varepsilon = \{\varepsilon_f, \varepsilon_{b_1}, \varepsilon_{b_2}, \varepsilon_{b_3}, \varepsilon_i, \varepsilon_{d_1}, \varepsilon_{d_2}, \varepsilon_{d_3}\}$$

**Step4**: Construct Gaussian probabilistic models with a mean determined by the output of PINNs and the adaptive weight collection $\varepsilon$.

**Step5**: Use $K$ gradient descent iterations to update the parameters $\varepsilon$ and $\theta$ as:

    **for** $j = 1$ to $K$ **do**

        (a) Resampling is performed every $K\omega$ iterations.

        (b) Define the adaptive weighted loss function.

        $L_{total}(\varepsilon; \theta; N)$ based on the maximum likelihood estimation.

(c) Tune the adaptive weight $\varepsilon$ via Adam optimizer to maximize the probability of meeting constraints.

$$\varepsilon_{j+1} \leftarrow Adam\left(L_{total}(\varepsilon_j;\theta_j;N)\right)$$

(d) Update the network weight $\theta$ via Adam optimizer.

$$\theta_{j+1} \leftarrow Adam\left(L_{total}(\varepsilon_j;\theta_j;N)\right)$$

**end for**

Return

The best model with parameters $\theta^*$ and the final adaptive weight $\varepsilon^*$.

## 4. Numerical experiments

In this section, we present four numerical experiments to verify the proposed new approach. Firstly, we aim to emphasize the ability of our method to solve 1D time-independent, 1D time-dependent, 2D time-dependent and 3D time-independent PNP equations. It is worth noting that we combine the self-adaptive loss balanced method and resampling strategy with the PINNs, enhancing the efficiency of PINNs in solving PNP equations. Subsequently, we compare the performance of EPINNs and PINNs. Unless explicitly stated otherwise, the default activation function is set to the hyperbolic tangent, and the resampling ratio is set to $\omega=0.4$. To evaluate the accuracy of the solutions, the following L$_2$ error is adopted:

$$L_2 \text{ error} = \frac{\sqrt{\sum_{j=1}^{N_{test}}\left|Y_{pred}(\mathbf{x}_{test}^j,t_{test}^j) - Y_{exact}(\mathbf{x}_{test}^j,t_{test}^j)\right|^2}}{\sqrt{\sum_{j=1}^{M}\left|Y_{exact}(\mathbf{x}_{test}^j,t_{test}^j)\right|^2}}, \tag{18}$$

where $Y_{pred}$ and $Y_{exact}$ represent numerical and exact solutions at test points $\{\mathbf{x}_{test}^j,t_{test}^j\}_{j=1}^{N_{test}}$, $N_{test}$ is the number of test points. In all experiments, the algorithm is implemented in Python 3.8 and TensorFlow 2.7.4 based on the personal computer with GPU GTX 1060.

### 4.1 One-dimensional time-independent PNP system

The first example aims to demonstrate the ability of EPINNs for solving 1D time-independent PNP equations. With this regard, we use the EPINNs to address a simple PNP system with Dirichlet boundary condition in the domain $\Omega=[-3,3]$. Meanwhile, the FDM and the PINNs are also employed to compare with our method. The problem is described as:

$$\begin{cases} \dfrac{\partial^2 c_p}{\partial x^2} = -\pi^2 (c_n + \phi), & x \in \Omega, \\ 3000\dfrac{\partial^2 c_n}{\partial x^2} + 100\left(\dfrac{\partial c_n}{\partial x}\dfrac{\partial c_p}{\partial x} + c_n \dfrac{\partial^2 c_p}{\partial x^2}\right) + f_v(x) = 0, & x \in \Omega, \\ 1000\dfrac{\partial^2 \phi}{\partial x^2} + 50\left(\dfrac{\partial \phi}{\partial x}\dfrac{\partial c_p}{\partial x} + \phi \dfrac{\partial^2 c_p}{\partial x^2}\right) + f_w(x) = 0, & x \in \Omega, \end{cases} \quad (19)$$

with boundary conditions:

$$\begin{cases} c_p(-3) = -1, \ c_p(3) = -1, \\ c_n(-3) = 0, \ c_n(3) = 0, \\ \phi(-3) = -1, \ \phi(3) = -1. \end{cases} \quad (20)$$

The analytical solutions for $c_p(x)$, $c_n(x)$ and $\phi(x)$ are separately defined as:

$$\begin{cases} c_p(x) = \sin(\pi x) + \cos(\pi x), \\ c_n(x) = \sin(\pi x), \\ \phi(x) = \cos(\pi x), \end{cases} \quad (21)$$

and the source terms $f_i(x)$ $(i = v, w)$ can be derived from the analytical solutions.

We investigate the above problem using the developed neural network with 4 layers and 15 neurons per hidden layer. The training data set was sampled randomly and consists of boundary training data $N_b = 2$, collocation points $N_f = 600$, true training data $N_{data} = 60$, test points data $N_{test} = 200$. The learning rate for the Adam optimizer is fixed at 0.001, the number of iterations is set to $K = 80000$, and the activation function is Tanh.

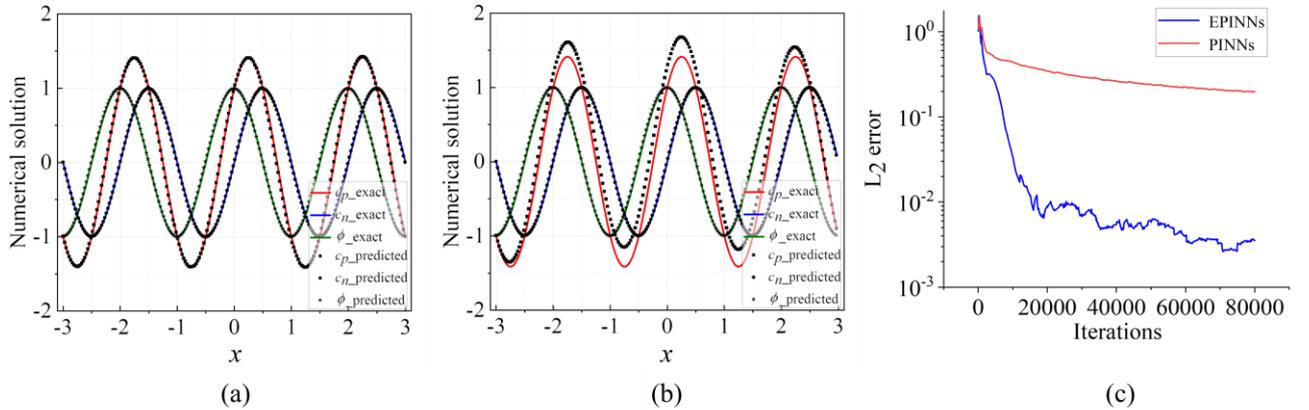

**Fig. 2.** Comparison of numerical results with exact solutions: (a) the EPINNs, (b) the PINNs, and (c) $L_2$ errors.

In Figs. 2(a) and 2(b), we compared the numerical results of both the traditional PINNs and the new EPINNs with the exact solutions. As can be seen, the results of our method are in good agreement with the exact solutions, while the traditional approach exhibits significant deviations. Fig. 2(c) illustrates the L$_2$ errors for both PINNs and EPINNs solutions, calculated as the average of five independent runs (unless otherwise specified, we opt for the L$_2$ errors averaged over five independent runs in the remaining numerical examples). It is noted that the computational accuracy of the new method surpasses that of the traditional approach by a considerable margin.

Next, we compare the developed EPINNs with the FDM and the traditional PINNs in Table 1. The error data in this table clearly indicates the superior computational accuracy of our method. This example provides preliminary evidence that the EPINNs can be used to address the PNP problem.

**Table 1.** The L$_2$ errors of obtained by using three different algorithms for solving one-dimensional time-independent PNP system.

| Algorithm | $c_p$ error | $c_n$ error | $\phi$ error |
|---|---|---|---|
| PINNs | $7.4\times10^{-1}$ | $1.2\times10^{-2}$ | $9.8\times10^{-3}$ |
| EPINNs | $9.5\times10^{-3}$ | $1.1\times10^{-3}$ | $9.7\times10^{-4}$ |
| FDMs | $5.0\times10^{-2}$ | $9.0\times10^{-3}$ | $9.0\times10^{-3}$ |

Keeping other parameters constant and setting the number of iterations to 40000, we examined the impact of the activation function on the accuracy of the proposed method. Three different activation functions (Tanh, Swish and Elu) in Table 2 were employed, and the resulting errors are depicted in Fig. 3. It is noticed that the Tanh function is relatively optimal, with an error of $3.8\times10^{-3}$. A significant number of experiments also indicate that the Tanh function is the optimal choice for solving the problem addressed in this paper; henceforth, we will default to using this function if not special instructions.

**Table 2.** Mathematical definitions of three different activation functions.

| Activation | Tanh | Swish | Elu |
|---|---|---|---|
| Formula | $\dfrac{e^x - e^{-x}}{e^x + e^{-x}}$ | $\dfrac{x}{1+e^{-x}}$ | $\begin{cases} x, x > 0 \\ a(e^x - 1), x \leq 0 \end{cases}$ |

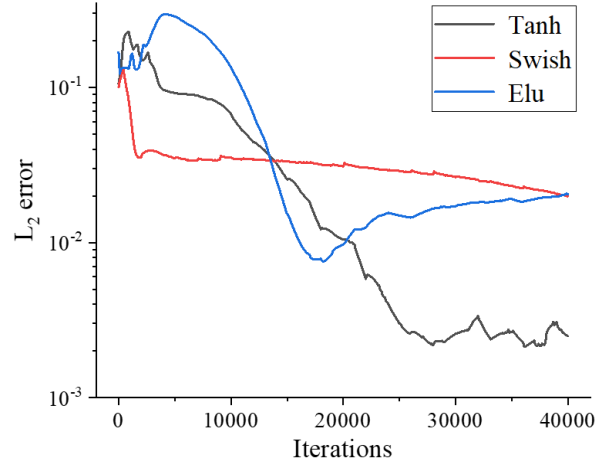

**Fig. 3.** The $L_2$ errors of the EPINNs with three different activation functions.

In addition, we also investigate the influence of the number of collocation points on the performance of the EPINNs. Table 3 lists the $L_2$ errors and computational times of the EPINNs under different numbers of collocation points. The overall trend indicates that, while keeping a constant neural network architecture, predictive accuracy shows an upward trend as the number of training collocation points increases. However, there was no significant increase in the computation time.

**Table 3.** $L_2$ errors and computational times of the EPINNs with different numbers of collocation points.

| $N_f$ | 200 | 400 | 600 | 800 | 1000 |
|---|---|---|---|---|---|
| $L_2$ errors | $1.5 \times 10^{-1}$ | $2.8 \times 10^{-2}$ | $1.3 \times 10^{-2}$ | $5.6 \times 10^{-3}$ | $3.9 \times 10^{-3}$ |
| Time (s) | 164.6 | 185.4 | 190.7 | 191.7 | 184.5 |

*4.2 One-dimensional time-dependent PNP system*

In this example, we consider a one-dimensional time-dependent PNP system in the domain $\Omega = [0,1]$, which is given by

$$\begin{cases} \dfrac{\partial c_p}{\partial t} = \dfrac{\partial^2 c_p}{\partial x^2} + q_1 \dfrac{\partial c_p}{\partial x} \dfrac{\partial \phi}{\partial x} + q_1 c_p \dfrac{\partial^2 \phi}{\partial x^2} + f_1, & (x,t) \in \Omega_t = \Omega \times [0,T], \\ \dfrac{\partial c_n}{\partial t} = \dfrac{\partial^2 c_n}{\partial x^2} + q_2 \dfrac{\partial c_n}{\partial x} \dfrac{\partial \phi}{\partial x} + q_2 c_n \dfrac{\partial^2 \phi}{\partial x^2} + f_2, & (x,t) \in \Omega_t = \Omega \times [0,T], \\ \dfrac{\partial^2 \phi}{\partial x^2} = -q_1 c_p - q_2 c_n, & (x,t) \in \Omega_t = \Omega \times [0,T], \end{cases} \quad (22)$$

with boundary conditions:

$$\begin{cases} \dfrac{\partial c_p}{\partial x}\bigg|_{x=0,1}=0, \\ \dfrac{\partial c_n}{\partial x}\bigg|_{x=0,1}=0, \\ \dfrac{\partial \phi}{\partial x}\bigg|_{x=0}=0,\ \dfrac{\partial \phi}{\partial x}\bigg|_{x=1}=-\dfrac{e^{-t}}{60}. \end{cases} \quad (23)$$

In above equations, $q_1=1$, $q_2=-1$ and $T=1$. $f_1$ and $f_2$ can be calculated by the following exact solution:

$$\begin{cases} c_p(x,t)=x^2(1-x)^2 e^{-t}, \\ c_n(x,t)=x^2(1-x)^3 e^{-t}, \\ \phi(x,t)=-(10x^7-28x^6+21x^5)e^{-t}/420. \end{cases} \quad (24)$$

In the calculation, we use a network structure with 6 layers and 24 neurons per hidden layer, and chose $N_f=7000$ collocation points from a random distribution. Besides, we fixed $N_b=500$, $N_i=400$, $N_{test}=1000$ and $N_{data}=150$. The learning rate for the Adam optimizer is set to 0.001, the activation function [29] is Tanh, and the number of iterations is set to 50000. To evaluate the accuracy of the method, we selected 70 test points uniformly distributed in the computational domain $\Omega_t$. At the same time, we conducted a comparative analysis between the original PINNs and the EPINNs under the same network structure.

Figs. 4 (a) and (b) shows the numerical solutions at $t=0.1$s obtained by the EPINNs and the PINNs, respectively. It is found that the proposed PINNs can effectively predict various physical quantities, while the traditional method falls short of expectations. As can be seen from Fig. 4 (b), there is a notable discrepancy between the PINN solutions and the true values for the electric potential $\phi$. After carefully examining loss values, we have identified a substantial disparity the electric potential $\phi$, leading to a significant increase in loss during the calculation process due to the order of magnitude difference. The EPINNs has effectively improved the original method by introducing adaptive weight techniques, enabling accurate and efficient simulation of such problems. The loss curves of the two methods are also provided in Fig. 4 (c). In the initial 10000 iterations, both PINNs and EPINNs exhibit similar error reduction capabilities. However, as the number of iterations increases, the advantages of EPINNs become more pronounced. The $L_2$ error of EPINNs solutions is smaller, indicating that EPINNs have a superior approximation ability for the PNP system.

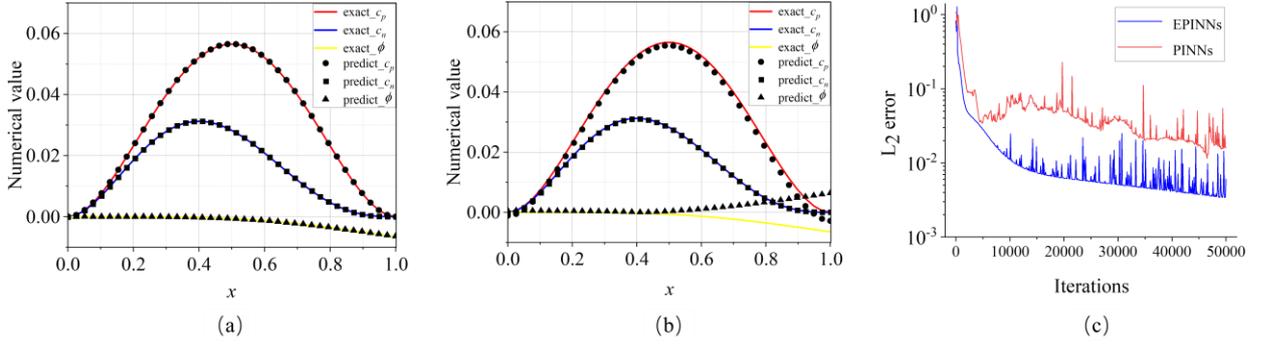

**Fig. 4.** Comparison of the EPINNs and PINNs for one-dimensional time-dependent PNP system: (a) the EPINNs, (b) the PINNs, and (c) loss curves.

The sampling method has a certain impact on the numerical results. Here, we compare three common sampling methods [30]: Pseudo random sampling, Uniform sampling, and Latin hypercube sampling (LHS). The distributions of points are illustrated in Fig. 5. The loss curves obtained using three different sampling methods are plotted in Fig. 6. It can be observed that the LHS has a comparative advantage over the other two methods, hence, we will employ this approach in subsequent calculations.

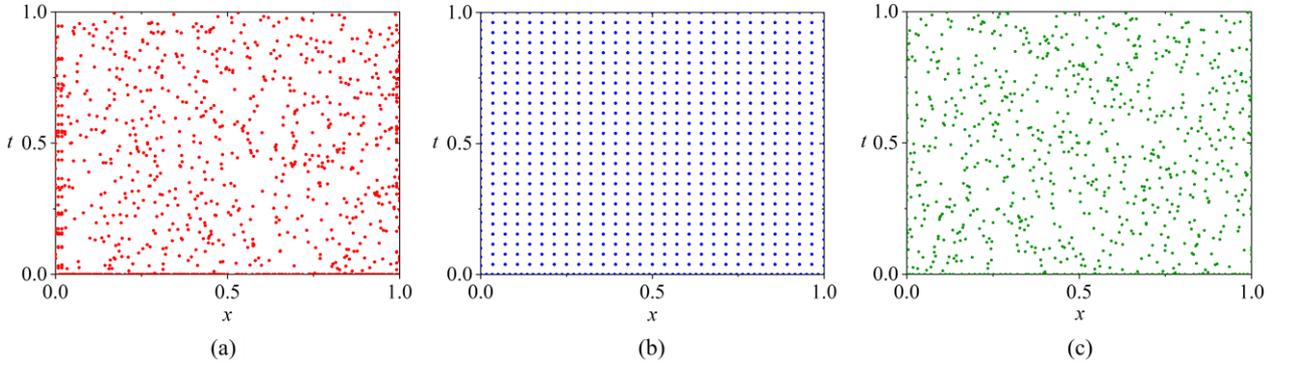

**Fig. 5.** Distribution of points for different sampling methods: (a) Pseudo-random sampling, (b) Uniform sampling, (c) Latin hypercube sampling.

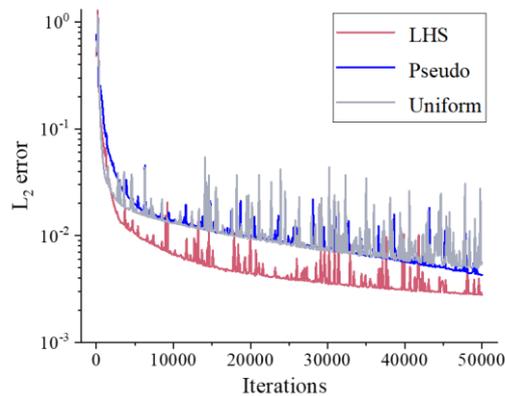

**Fig. 6.** The loss curves of the EPINNs for one-dimensional time-dependent PNP system by using three different sampling methods: LHS, Pseudo and Uniform.

In addition, Fig. 7 gives the snapshots of $c_p$, $c_n$ and $\phi$ at different time stages $t=0\text{s}$, $t=0.2\text{s}$, $t=0.4\text{s}$, $t=0.6\text{s}$, $t=0.8\text{s}$, and $t=1\text{s}$. Clearly, the absolute values of the outputs decrease over time, exhibiting a regularity consistent with Eq. (24). This conformity reinforces the reliability and accuracy of the EPINNs in capturing the underlying behavior of the Eq. (22). Furthermore, it can be seen from Table 4 that the $L_2$ errors of the method remains basically unchanged at different times, indicating the accuracy and stability of the proposed method in dealing with such problems.

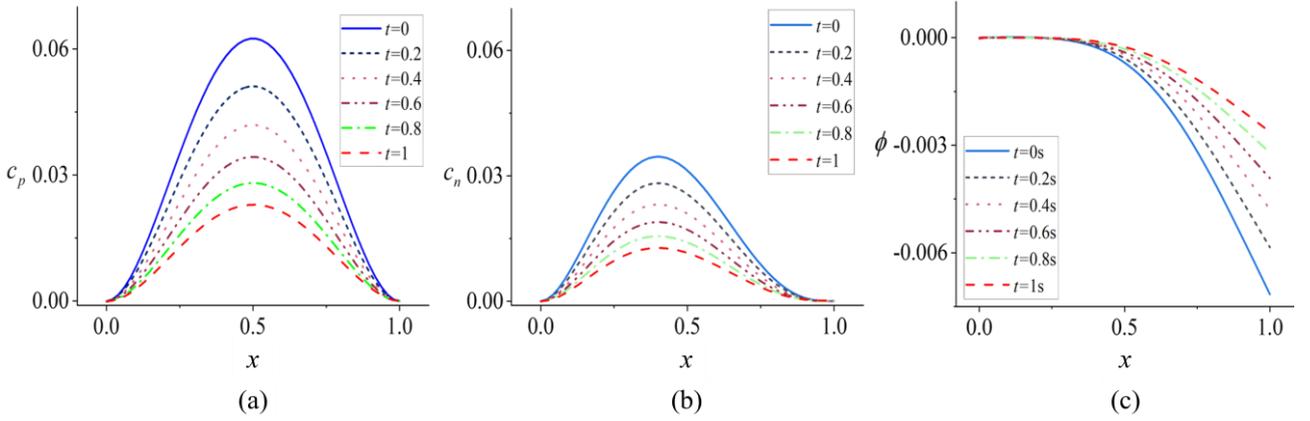

**Fig. 7.** Snapshots of the EPINNs approximation at various time points for one-dimensional time-dependent PNP system: (a) $c_p$, (b) $c_n$ and (c) $\phi$.

Table 4. $L_2$ error of the proposed EPINNs at different time points.

| $t$ (s) | $c_p$ error | $c_n$ error | $\phi$ error |
| --- | --- | --- | --- |
| 0 | $3.5\times10^{-4}$ | $2.7\times10^{-4}$ | $3.1\times10^{-3}$ |
| 0.2 | $1.3\times10^{-4}$ | $2.5\times10^{-4}$ | $4.4\times10^{-3}$ |
| 0.4 | $3.0\times10^{-4}$ | $6.1\times10^{-4}$ | $5.6\times10^{-3}$ |
| 0.6 | $2.8\times10^{-4}$ | $1.2\times10^{-4}$ | $2.7\times10^{-3}$ |
| 0.8 | $4.9\times10^{-4}$ | $1.2\times10^{-4}$ | $3.3\times10^{-3}$ |
| 1 | $9.8\times10^{-5}$ | $1.8\times10^{-4}$ | $4.0\times10^{-3}$ |

Finally, keeping $N_b=400$, $N_i=400$, $N_f=7000$, $N_{test}=1000$, $N_{data}=150$ and $K=30000$ constant, we examined the impact of neural network size on the computational results. Table 5 presents the average $L_2$ errors from 10 runs for various numbers of hidden layers and neurons per layer. We observe that both the number of neurons and the number of hidden layers have minimal impact on computational accuracy.

Table 5. $L_2$ errors of the EPINNs under different network sizes.

| Neurons / Layers | 10 | 20 | 40 |
|---|---|---|---|
| 2 | $1.7\times10^{-2}$ | $3.7\times10^{-3}$ | $3.6\times10^{-2}$ |
| 4 | $2.0\times10^{-2}$ | $8.1\times10^{-3}$ | $1.0\times10^{-2}$ |
| 6 | $3.0\times10^{-2}$ | $1.5\times10^{-3}$ | $4.0\times10^{-2}$ |
| 8 | $1.1\times10^{-2}$ | $3.2\times10^{-3}$ | $6.7\times10^{-2}$ |

*4.3 Two-dimensional time-dependent PNP equations*

In the third example, we investigate the following 2D time-dependent PNP system in $\Omega=[0,1]^2$ with boundary $\partial\Omega$ [31]:

$$\begin{cases} \dfrac{\partial c_p}{\partial t}-\nabla\cdot\left(\nabla c_p+c_p\nabla\phi\right)=f_1, & (\mathbf{x},t)\in\Omega\times(0,T], \\ \dfrac{\partial c_n}{\partial t}-\nabla\cdot\left(\nabla c_n-c_n\nabla\phi\right)=f_2, & (\mathbf{x},t)\in\Omega\times(0,T], \\ -\Delta\phi-c_p+c_n=f_3, & (\mathbf{x},t)\in\Omega\times(0,T], \\ c_p=c_n=\phi=0, & (\mathbf{x},t)\in\partial\Omega\times(0,T], \\ c_p(\mathbf{x},0)=c_p^0(\mathbf{x}),\ c_n(\mathbf{x},0)=c_n^0(\mathbf{x}), & \mathbf{x}\in\Omega, \end{cases} \quad (25)$$

where $\mathbf{x}=\{x,y\}$, $T=1$, $f_i(i=1,2,3)$ are the reaction terms, $c_p^0$ and $c_n^0$ are known functions. The exact solutions of the above problem are

$$\begin{cases} c_p(x,y,t)=\sin(2\pi x)\sin(2\pi y)\sin(t), \\ c_n(x,y,t)=\sin(3\pi x)\sin(3\pi y)\sin(2t), \\ \phi(x,y,t)=\sin(\pi x)\sin(\pi y)\left(1-e^{-t}\right). \end{cases} \quad (26)$$

In the computation, we use a neural network with 6 layers and 25 neurons per hidden layer, and select the Tanh function as activation function. Additionally, we set $N_b=400$, $N_f=8000$, $N_i=200$, $N_{data}=1000$ and $N_{test}=10000$, the iteration is set to $K=30000$, The learning rate for the Adam optimizer is set to 0.001, and all the training data set were sampled by LHS.

Figs. 8, 9 and 10 show the profiles of $c_p$, $c_n$ and $\phi$ in the computational domain at $t=0.1\mathrm{s}$, respectively. Each figure includes exact solutions, PINN solutions, and EPINN solutions. From these figures, it is evident that the approximation capability of PINNs is significantly limited. Conversely, the EPINNs remains robust in addressing this problem, reaffirming the superiority of EPINNs in solving PNP equations. Furthermore, Fig. 11 provides the error distributions of the EPINNs for $c_p$, $c_n$ and $\phi$, the maximum error does not exceed $3.0\times10^{-4}$. These numerical experiments demonstrate that the present EPINNs is an accurate, stable, and effective approach for solving PNP systems.

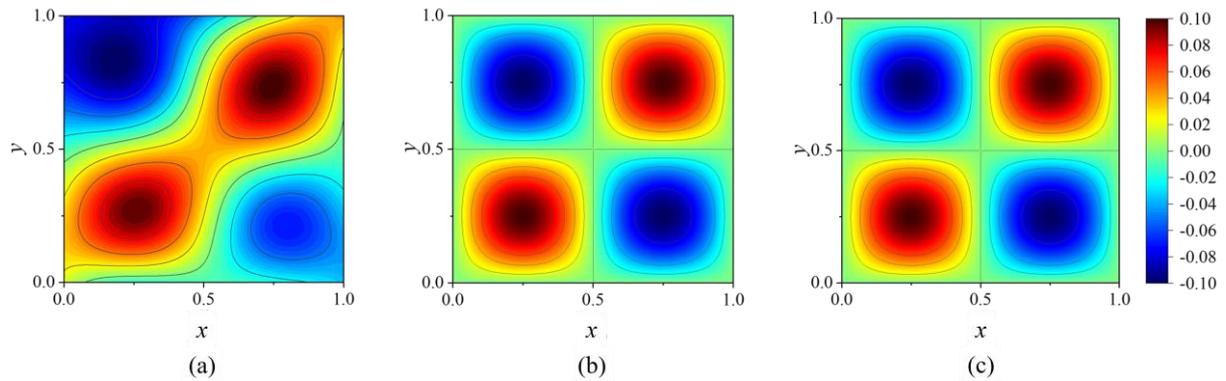

**Fig. 8.** Comparison between numerical results and exact values for $c_p$ at $t = 0.1$s : (a) PINNs solution, (b) EPINNs solution, (c) exact solution.

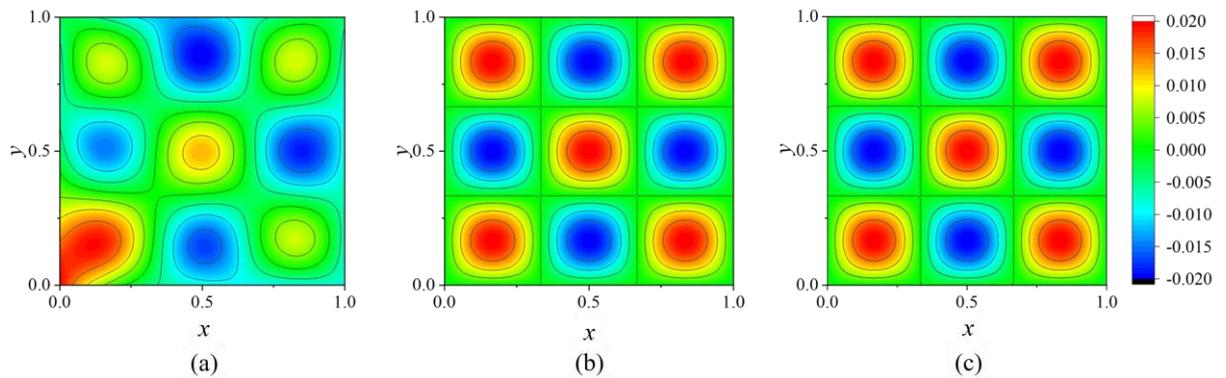

**Fig. 9.** Comparison between numerical results and exact values for $c_n$ at $t = 0.1$s : (a) PINNs solution, (b) EPINNs solution, (c) exact solution.

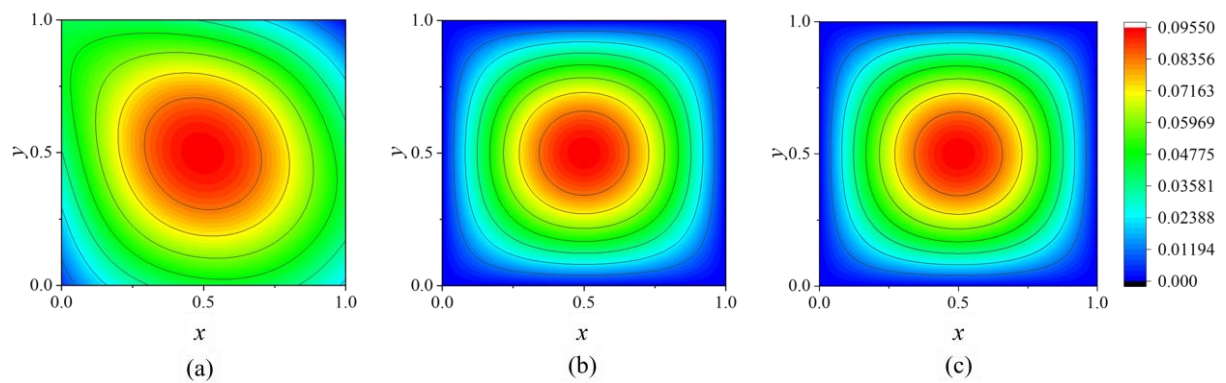

Fig. 10. Comparison between numerical results and exact values for $\phi$ at $t = 0.1\text{s}$: (a) PINNs solution, (b) EPINNs solution, (c) exact solution.

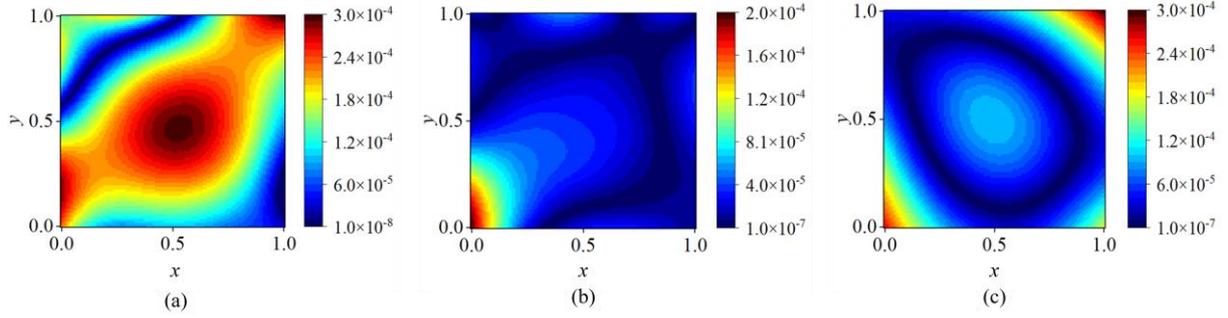

Fig. 11. The $L_2$ errors of the EPINNs: (a) $c_p$, (b) $c_n$, (c) $\phi$.

To further assess the accuracy of EPINNs in addressing two-dimensional time-dependent PNP problems, we employed the EPINNs to predict snapshots of $c_p$, $c_n$ and $\phi$ at different time points ($t = 0.25\text{s}, 0.50\text{s}, 0.75\text{s}, 1\text{s}$). Figs. 12, 13 and 14 respectively illustrate the predicted results for $c_p$, $c_n$ and $\phi$. From these figures, it can be observed that the distribution of concentration and electric potential becomes increasingly uneven with the passage of time. The corresponding $L_2$ errors are also listed in Table 6. We can noticed that the computational error of the method does not escalate with the passage of time. This once again confirms the accuracy and effectiveness of our method.

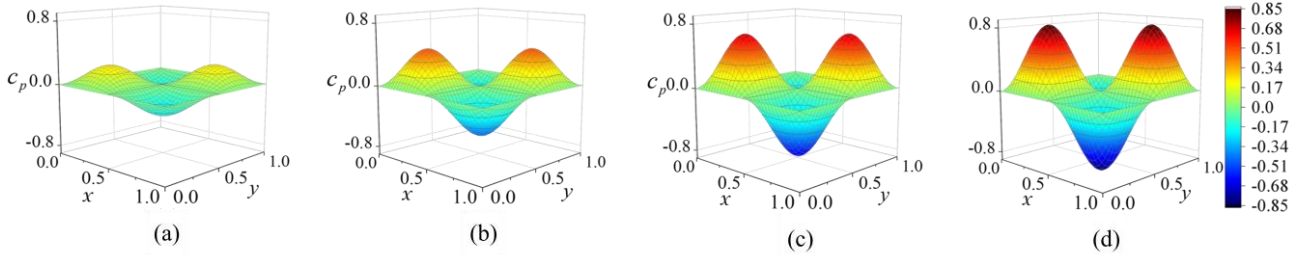

Fig. 12. Profiles of concentration $c_p$ at different time points predicted by the EPINNs: (a) $t = 0.25\text{s}$, (b) $t = 0.50\text{s}$, (c) $t = 0.75\text{s}$, (d) $t = 1.00\text{s}$.

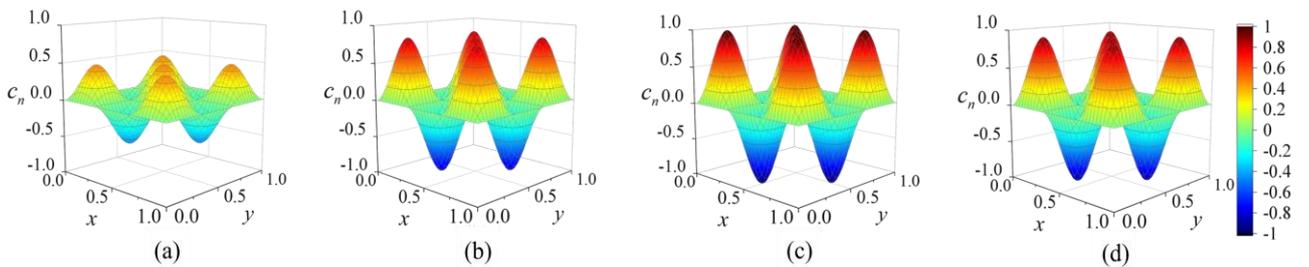

**Fig. 13.** Profiles of concentration $c_n$ at different time points predicted by the EPINNs: (a) $t = 0.25\text{s}$, (b) $t = 0.50\text{s}$, (c) $t = 0.75\text{s}$, (d) $t = 1.00\text{s}$.

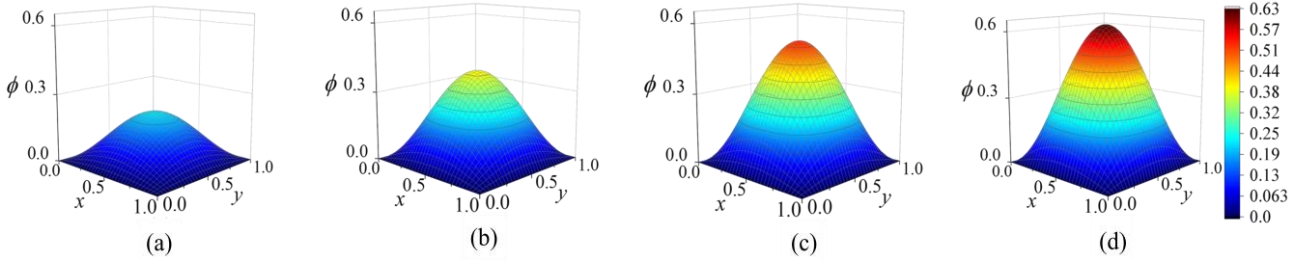

(a)     (b)     (c)     (d)

**Fig. 14.** Profiles of electric potential $\phi$ at different time points predicted by the EPINNs: (a) $t = 0.25\text{s}$, (b) $t = 0.50\text{s}$, (c) $t = 0.75\text{s}$, (d) $t = 1.00\text{s}$.

**Table 6.** The $L_2$ errors of numerical results at different time points.

| Time/s | $c_p$ error | $c_n$ error | $\phi$ error |
|---|---|---|---|
| 0.25 | $5.0 \times 10^{-3}$ | $4.9 \times 10^{-3}$ | $4.7 \times 10^{-3}$ |
| 0.50 | $4.2 \times 10^{-3}$ | $3.7 \times 10^{-3}$ | $2.4 \times 10^{-3}$ |
| 0.75 | $4.4 \times 10^{-3}$ | $4.2 \times 10^{-3}$ | $4.0 \times 10^{-3}$ |
| 1.00 | $4.6 \times 10^{-3}$ | $3.1 \times 10^{-3}$ | $4.3 \times 10^{-3}$ |

*4.4 Three-dimensional time-independent PNP equations*

In the last example, the proposed scheme is applied to the 3D-time-independent PNP problem [32]. We consider a dimensionless PNP system with source terms in a domain $\Omega = [-\frac{L}{2}, \frac{L}{2}]^3$, the governing equations are given by:

$$\begin{cases} -\nabla \cdot (\varepsilon \nabla \phi) = \dfrac{e_c^2 \beta}{\varepsilon_0}(c_{K^+} - c_{Cl^-}), & (x, y, z) \in \Omega, \\ \nabla \cdot D_{K^+}(\nabla c_{K^+} + z_{K^+} c_{K^+} \nabla \phi) + f_{K^+} = 0, & (x, y, z) \in \Omega, \\ \nabla \cdot D_{Cl^-}(\nabla c_{Cl^-} + z_{Cl^-} c_{Cl^-} \nabla \phi) + f_{Cl^-} = 0, & (x, y, z) \in \Omega, \end{cases} \qquad (27)$$

where $\varepsilon_0$ is the permittivity of vacuum, $\varepsilon = 80$ is the dielectric constant, $L = 2 \times 10^{-7}\text{m}$, $z_{K^+} = 1$, $z_{Cl^-} = -1$. The diffusion coefficients are $D_{K^+} = 1.96 \times 10^{-9}\text{m}^2/\text{s}$ and $D_{Cl^-} = 2.03 \times 10^{-9}\text{m}^2/\text{s}$. The boundary conditions are given by

$$\begin{cases} \phi(x,y,z) = e_c \beta V_{app} \dfrac{z+L/2}{L}, & (x,y,z) \in \partial\Omega \\ c_{K^+}(x,y,z) = c^b, & (x,y,z) \in \partial\Omega \\ c_{Cl^-}(x,y,z) = c^b, & (x,y,z) \in \partial\Omega \end{cases} \quad (28)$$

where $c^b = 5\times 10^{-4}$ m, and $\partial\Omega$ is the boundary of domain $\Omega$. The exact solutions for $c_{K^+}$, $c_{Cl^-}$ and $\phi$ are given by:

$$\begin{cases} c_{K^+} = c^b + \dfrac{c^b}{2}\cos(\dfrac{\pi x}{L})\cos(\dfrac{\pi y}{L})\cos(\dfrac{\pi z}{L}), \\ c_{Cl^-} = c^b - \dfrac{c^b}{2}\cos(\dfrac{\pi x}{L})\cos(\dfrac{\pi y}{L})\cos(\dfrac{\pi z}{L}), \\ \phi = \tau \cos(\dfrac{\pi x}{L})\cos(\dfrac{\pi y}{L})\cos(\dfrac{\pi z}{L}) + e_c \beta V_{app} \dfrac{z+L/2}{L}, \end{cases} \quad (29)$$

where $\tau = \dfrac{L^2}{3\pi^2 \varepsilon} c^b \dfrac{e_c^2 \beta}{\varepsilon_0} = 3.583\times 10^{-7}$, and $V_{app}$ represents applied voltage.

A neural network with 6 layers and 25 neurons per hidden layer is established to address this 3D PNP system. The Tanh is adopted as activation function and learning rate is set to 0.001 in this example. We set $N_b = 800$, $N_f = 10000$, $N_{test} = 10000$ and $N_{data} = 1500$, and the iterations is set to $K = 20000$, all the training data set were sampled by LHS.

The EPINNs framework is employed to estimate the spatial distribution of electrostatic potential on the left surface of the computational domain under various applied voltages. Fig. 15 shows numerical results of electrostatic potential with three different voltages ($V_{app} = 0\text{V}$, $V_{app} = 1\text{V}$, and $V_{app} = 3\text{V}$), and corresponding exact solutions are depicted in Fig. 16. It can be seen from Figs. 15 and 16 that the numerical results are in good agreement with the exact solutions. It is worth noting that the geometric dimensions of the case are at the micron level, while the numerical values of the electrostatic potential are close to $10^{-17}$. Even so, our method is still can accurately and effectively simulate such problems. This validates the performance and applicability of the proposed approach for solving three-dimensional PNP systems.

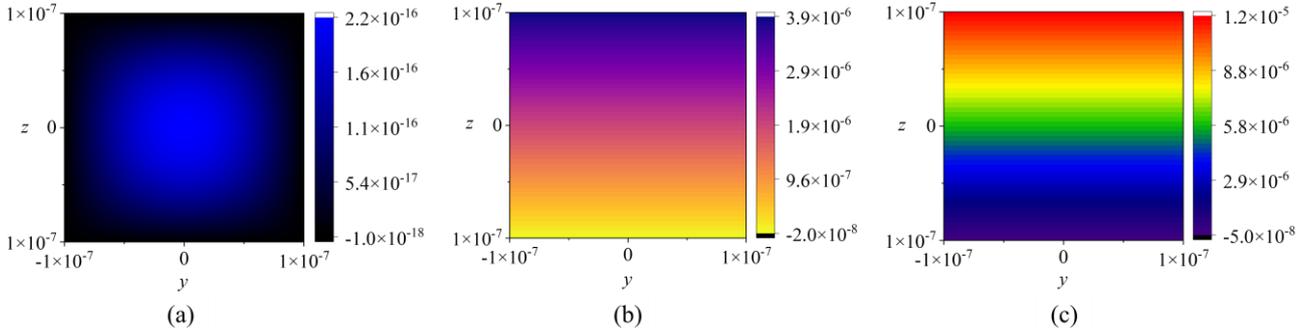

**Fig. 15.** Numerical distributions of electrostatic potential on the left surface of the computational domain under different voltage environments: (a) $V_{app} = 0\text{V}$, (b) $V_{app} = 1\text{V}$, (c) $V_{app} = 3\text{V}$.

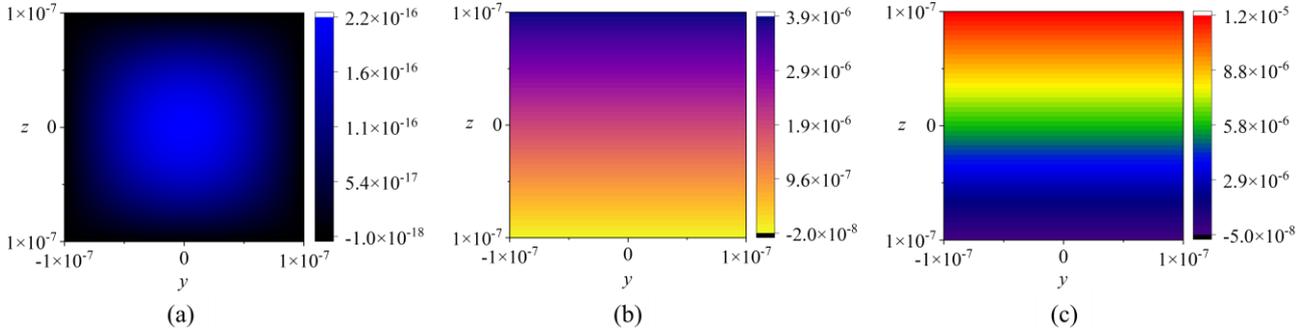

**Fig. 16.** Exact distributions of electrostatic potential on the left surface of the computational domain under different voltage environments: (a) $V_{app} = 0\text{V}$, (b) $V_{app} = 1\text{V}$, (c) $V_{app} = 3\text{V}$.

## 5. Conclusion

In this paper, we present an enriched physical-informed neural networks (EPINNs) for solving PNP equations in various dimensions (including 1D, 2D and 3D). The conventional PINNs have proven ineffective in addressing PNP problem due to its nonlinear and strongly coupled characteristics. By incorporating an adaptive loss weights technique into the original PINNs, the imbalance issues between different sub-losses are effectively mitigated. This enhancement enables the new neural networks to accurately and efficiently simulate PNP systems. Furthermore, the use of resampling method further improves the computational accuracy of the proposed approach.

Four numerical experiments, including time-independent and time-dependent PNP systems, are presented to investigate the performance of the proposed EPINNs. Numerical results indicate that the new method can accurately and stably simulate various PNP systems. Even when dealing with geometric dimensions at the micrometer scale and exceedingly small values of physical quantities, the method retains its precision and effectiveness. Furthermore, our approach does not lose computational accuracy in solving such dynamic problems over time. In summary, this article provides a simple and accurate intelligent algorithm for solving PNP systems, which could be extended to simulate electrochemical ion transport processes.

The method developed in this article belongs to meshless methods, and does not require grid generation and numerical integration. For time-dependent problems, the method directly discretize the spatio-temporal domain instead of difference methods, making it more time-consuming for high-dimensional problems. Enhancing the efficiency in addressing prolonged and large-scale dynamic problems will be the focal point of our future endeavors.

**Acknowledgements**

The work described in this paper was supported by the Natural Science Foundation of Shandong Province of China (Grant No. ZR2023YQ005). Dr. Fajie Wang gratefully acknowledges the support of K. C. Wong Education Foundation and DAAD.

**Data availability statement**

The data that support the findings of this study are available from the corresponding author upon reasonable request.

**Conflicts of Interest**

The authors declare no conflict of interest.